\pdfoutput=1
\documentclass[10pt,twocolumn,letterpaper]{article}

\usepackage{cvpr}              % To produce the CAMERA-READY version
\usepackage{graphicx}
\usepackage{amsmath}
\usepackage{amssymb}
\usepackage{booktabs}
\usepackage{url}
\usepackage{color}
\usepackage{multirow}
\usepackage{pifont}
\usepackage[table]{xcolor}
\usepackage[group-separator={,}]{siunitx}

\usepackage[pagebackref,breaklinks,colorlinks]{hyperref}

\usepackage[capitalize]{cleveref}
\crefname{section}{Sec.}{Secs.}
\Crefname{section}{Section}{Sections}
\Crefname{table}{Table}{Tables}
\crefname{table}{Tab.}{Tabs.}

\def\cvprPaperID{7724} % *** Enter the CVPR Paper ID here
\def\confName{CVPR}
\def\confYear{2022}

\begin{document}

\title{BigDetection: A Large-scale Benchmark for\\
Improved Object Detector Pre-training}

\author{
Likun Cai$^{1}$\thanks{Work done during an internship at Amazon.} \quad Zhi Zhang$^{2}$ \quad Yi Zhu$^{2}$ \quad Li Zhang$^{1}$ \quad Mu Li$^{2}$ \quad Xiangyang Xue$^{1}$ \\
$^{1}$Fudan University \qquad $^{2}$Amazon Inc. \qquad\\
% {\tt\small\{lkcai20, \}.fudan.edu.cn }
% {\tt\small cddlyf@gmail.com }\\
}

\maketitle

%%%%%%%%% ABSTRACT

\begin{abstract}
Multiple datasets and open challenges for object detection have been introduced in recent years. 
To build more general and powerful object detection systems, in this paper, we construct a new large-scale benchmark termed BigDetection. 
Our goal is to simply leverage the training data from existing datasets (LVIS, OpenImages and Object365) with carefully designed principles, and curate a larger dataset for improved detector pre-training.
Specifically, we generate a new taxonomy which unifies the heterogeneous label spaces from different sources. 
Our BigDetection dataset has 600 object categories and contains over 3.4M training images with 36M bounding boxes. 
It is much larger in multiple dimensions than previous benchmarks, which offers both opportunities and challenges.
Extensive experiments demonstrate its validity as a new benchmark for evaluating different object detection methods, and its effectiveness as a pre-training dataset.
The code and models are available at \url{https://github.com/amazon-research/bigdetection}.

\end{abstract}

%%%%%%%%% BODY TEXT

\section{Introduction}
\label{sec:intro}

Back in 2014, Microsoft COCO dataset~\cite{lin2014microsoft} was an extremely challenging benchmark where best performing methods were claiming average precision scores less than 20 AP across all 80 categories.
Now, state-of-the-art detectors~\cite{xu2021end,dai2021dynamic} are already able to achieve 60+ AP on COCO test-dev.
As a golden standard, COCO has incubated many popular object detection algorithms.

\begin{figure}[t]
    \centering
    \fbox{\includegraphics[width=0.45\linewidth]{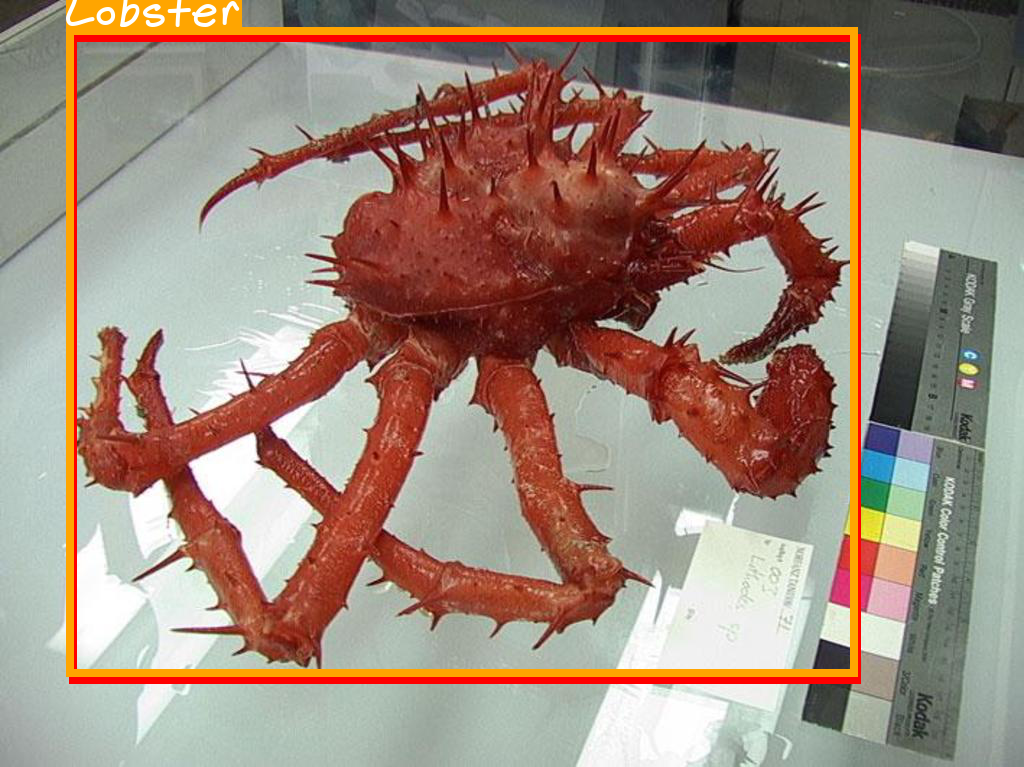}}
    \fbox{\includegraphics[width=0.45\linewidth]{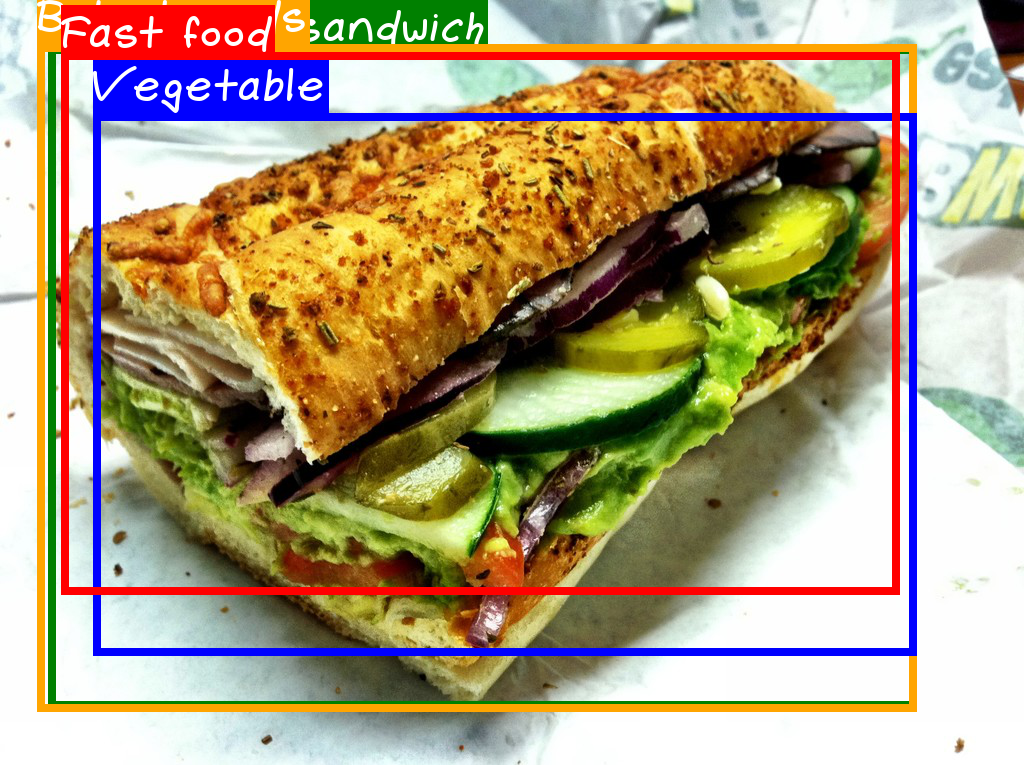}}
    \fbox{\includegraphics[width=0.45\linewidth]{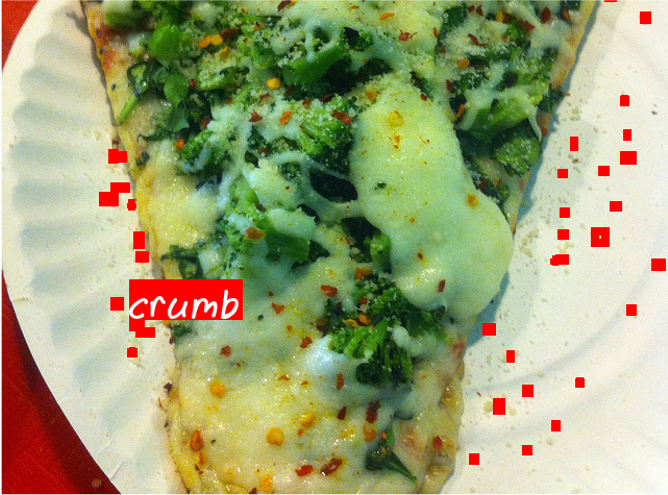}}
    \fbox{\includegraphics[width=0.45\linewidth]{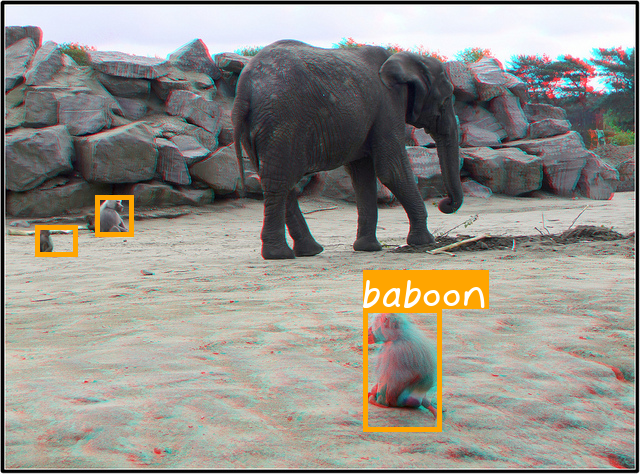}}
    
    \caption{Visual examples from OpenImages (top) and LVIS (bottom) datasets. Top left (wrong label): ``\textit{Crab}'' mistakenly labeled as ``\textit{Lobster}''. Top right (bbox overlapping): bboxes with different class labels locate at the same place.  Bottom left (uninformative annotations): class ``\textit{crumb}'' may not be useful for detector pre-training. Bottom right (long-tail): there is only one image with ``\textit{baboon}'' in the dataset.}
    \label{fig:lvis_oid_badcase}
    \vspace{-2ex}
\end{figure}

To build more robust and general object detection systems, several larger-scale object detection datasets have been released, such as OpenImages~\cite{kuznetsova2020open}, Objects365~\cite{shao2019objects365}, and LVIS~\cite{gupta2019lvis}.
However, each dataset has its own limitations and challenges. 
For example, OpenImages has around $10\%$ bounding box annotations that are machine-generated, which may cause problems like wrong label and bounding box overlapping  (Fig.~\ref{fig:lvis_oid_badcase} top).
LVIS aims to craft a diverse set of densely annotated labels covering more than 1200 categories, but may bring problems like uninformative annotation and serious long-tail distribution (Fig.~\ref{fig:lvis_oid_badcase} bottom). 
Object365 has a relatively smaller vocabulary which may miss common object categories like insect.

\begin{table*}[t]
    \begin{center}
        \begin{tabular}{l|cc|cc|c}
            \toprule
            ~ & \multicolumn{2}{c|}{Train} & \multicolumn{2}{c|}{Val} & \multirow{2}*{Num. classes} \\
            ~ & Num. images & Num. boxes & Num. images & Num. boxes & ~  \\
            \midrule
            % LVIS & \num{100170} & \num{1270141} & \num{19809} & \num{244707} & \num{1203} \\
            LVIS~\cite{gupta2019lvis} & 100K & 1.27M & 19K & 244K & 1203 \\
            % OpenImages & \num{1743042} & \num{14610229} & \num{41620} & \num{303980} & 600 \\
            OpenImages~\cite{kuznetsova2020open} & 1.74M & 14.61M & 41K & 303K & 600 \\
            % Objects365 & \num{1728775} & \num{22886232} & \num{80000} & \num{1062238} & 365 \\
            Objects365~\cite{shao2019objects365} & 1.72M & 22.89M & 80K & 1.06M & 365 \\
            % BigDetection & \num{3487533} & \num{35962268} & \num{141429} & \num{1587792} & 600 \\ 
            \midrule
            \textbf{BigDetection} & 3.48M & 35.96M & 141K & 1.58M & 600 \\
            \bottomrule
        \end{tabular}
    \end{center}
    \vspace{-2ex}
    \caption{Comparison of the dataset statistics among popular large-scale object detection benchmarks.}
    \label{tab:dataset}
\end{table*}

In this work, we introduce a new large-scale object detection benchmark, termed \textit{BigDetection}. Our goal is to simply leverage the training data from existing datasets (like LVIS, OpenImages and Objects365)  with carefully designed principles, so that we can curate a larger dataset more suitable for object detector pre-training.
Different from literature in multi-dataset detector training~\cite{wang2019towards, zhou2021simple, zhao2020object}, we use language model to build our initial unified label space across datasets and perform manual verification to obtain the final taxonomy. 
Our BigDetection dataset has 600 object categories and contains 3.4M training images with 36M bounding boxes. We show the statistics comparison to other datasets in~\cref{tab:dataset}.
In addition, we perform various experiments to demonstrate its validity as a new benchmark for evaluating different object detection methods, and its effectiveness as a pre-training dataset.
In particular, as we can see in~\cref{tab:swin}, a CBNetv2~\cite{liang2021cbnetv2} model with Swin-Base backbone~\cite{liu2021swin} pre-trained on BigDetection can achieve $59.8$ AP on COCO test-dev set. It is surprising to find that this performance is competitive with the same model using Swin-Large backbone without pre-training on BigDetection. Note that Swin-Large is twice heavier than the Swin-Base model.
In addition, following a partially labeled data setting~\cite{sohn2020simple} on COCO, BigDetection pre-training is shown to be extremely data efficient. \eg, $25.3$ AP on COCO validation set using only $1\%$ COCO training data.

\begin{figure}[t]
  \centering
  \includegraphics[width=0.8\linewidth]{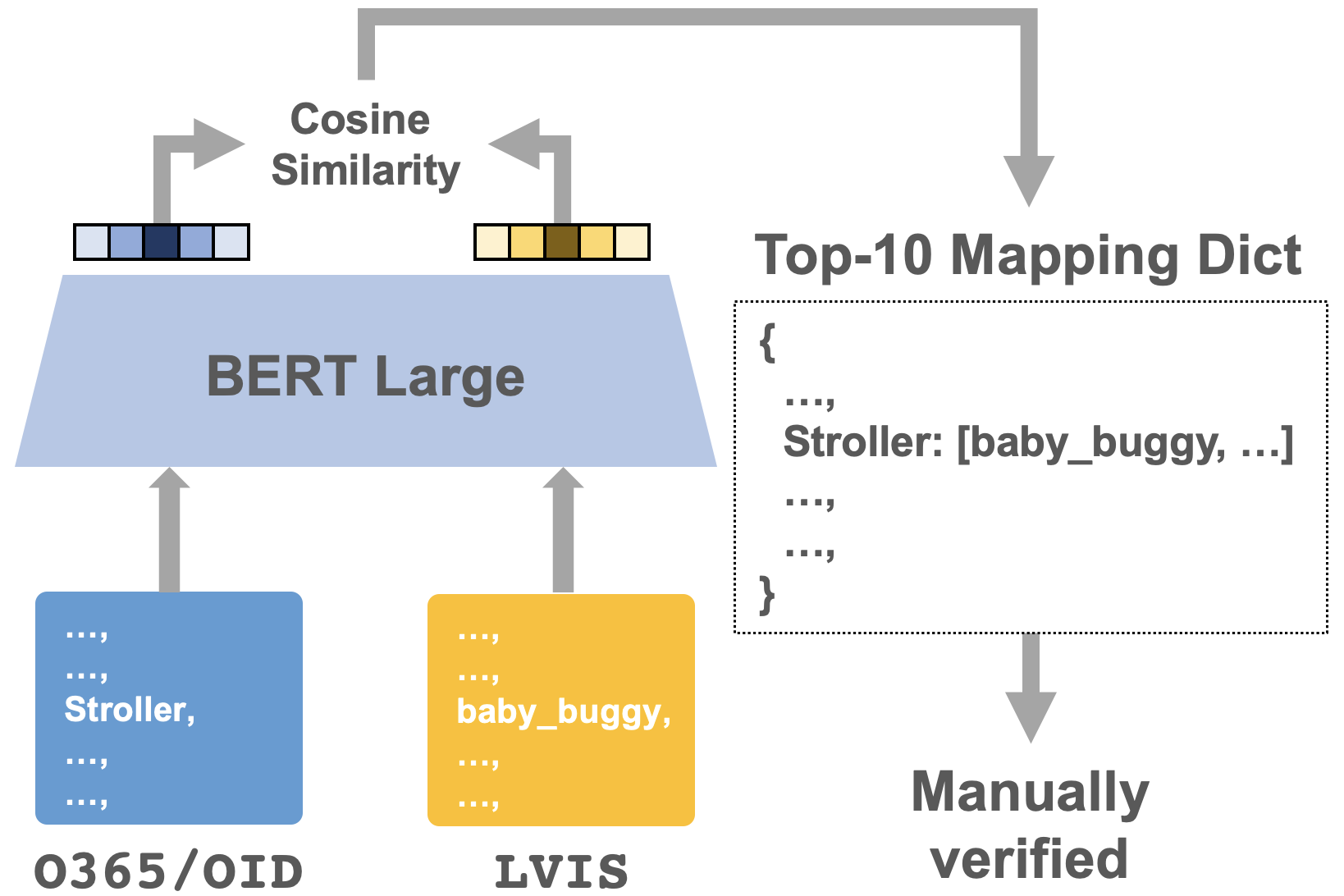}
  \caption{Overview of our category mapping pipeline, which is used to generate the unified label space of BigDetection. See text in Sec.~\ref{subsec:build} for  more details.}
  \label{fig:cat_mapping}
  \vspace{-2ex}
\end{figure}
Our contributions can be summarized as follows:
\begin{itemize}
    \item We introduce a new object detection dataset, BigDetection, which is much larger in multiple dimensions than previous benchmarks. It could serve as a more challenging benchmark for evaluating different object detection methods.
    \item We show effectiveness of BigDetection as a pre-training dataset. We obtain state-of-the-art results on COCO validation and test-dev sets, as well as under data-efficiency settings.
    \item We perform extensive ablation studies to provide good practices when training object detectors on large-scale datasets. 
\end{itemize}

\section{Related Work}
\label{sec:related}
% \noindent \textbf{Datasets for object detection}
\paragraph{Datasets for object detection}
Large-scale datasets with high-quality annotations play a crucial role in advancement of better computer vision models. 
In terms of object detection, \textsc{Pascal} VOC~\cite{everingham2015pascal} is one early benchmark containing $20$ classes over $17$k images.
Despite its relatively small scale compared to datasets nowadays, \textsc{Pascal} VOC has successfully bred many object detectors including both classical detectors~\cite{felzenszwalb2009object, wang2013regionlets} and deep learning detectors~\cite{girshick2014rich, he2015spatial, girshick2015fast}. 
Then comes Microsoft COCO~\cite{lin2014microsoft} in year 2014, which is the most widely adopted benchmark for object detection to present.
It contains $118$k images and $860$k instance annotations over $80$ classes.
Thanks to its large-scale and great quality, COCO together with deep learning have revolutionized the landscape of computer vision.
Recently, with extensive high quality labeling efforts, larger scale datasets like LVIS~\cite{gupta2019lvis}, OpenImages~\cite{kuznetsova2020open} and Objects365~\cite{shao2019objects365} are introduced with millions of instance-level annotations. 
They enable us to learn diversified and fine-grained object concepts, as well as explore the possibility of few-/zero-shot learning on new scenes. 
There are also more datasets for object detection in specific domains, such as~\cite{geiger2012we,cordts2016cityscapes,sun2020scalability,shao2018crowdhuman,real2017youtube,van2018inaturalist}, to support various use cases.

% \noindent \textbf{Multi-dataset detector training}
\paragraph{Multi-dataset detector training}
Annotating gigantic datasets by human labor is not scalable. Hence some recent work start to explore multi-dataset training strategy, whose goal is to learn better feature representations from more labeled data given existing datasets. 

One early attempt~\cite{wang2019towards} proposes to train a universal object detector with domain attention on multiple datasets. All parameters and computations are shared so that one detector can leverage knowledge across domains. 
In order to address the partial annotation problem when using multiple datasets with heterogeneous label space, UOD~\cite{zhao2020object} exploits a pseudo labeling mechanism to unify the label space for training a single detector.
Following~\cite{wang2019towards}, Zhou et al.~\cite{zhou2021simple} proposes a weighted graph matching behind split classifiers to automatically generate a common taxonomy. This framework can generalize better to new test domains without prior knowledge, and achieves great zero-shot performance.
In order to alleviate the scale variation problem, USB~\cite{shinya2021usb} introduces a universal-scale object detection benchmark to enable multi-scale object detection. Their proposed UniverseNet achieves top performance on two challenges. 

Different from the above work, we construct the final unified label space through a carefully designed mapping pipeline and strict manual verification, which makes our unified label space more credible than those machine generated results. In addition, existing object detectors can be trained directly on our dataset without any modifications like split classifiers, domain attention or graph matching. Thus our composite dataset provides a new benchmark that easily enables fair comparison. There is a recent work, MSeg~\cite{lambert2020mseg}, that is similar to us in terms of manually building a composite dataset. However, MSeg is designed for semantic segmentation and it only contains 200k images over 194 semantics classes. Our composite dataset is significantly larger, and we provide both clear benefits of pre-training and large-scale analysis.

\begin{figure}[t]
    \centering
    \subfloat[Before]{
        \includegraphics[width=0.45\linewidth]{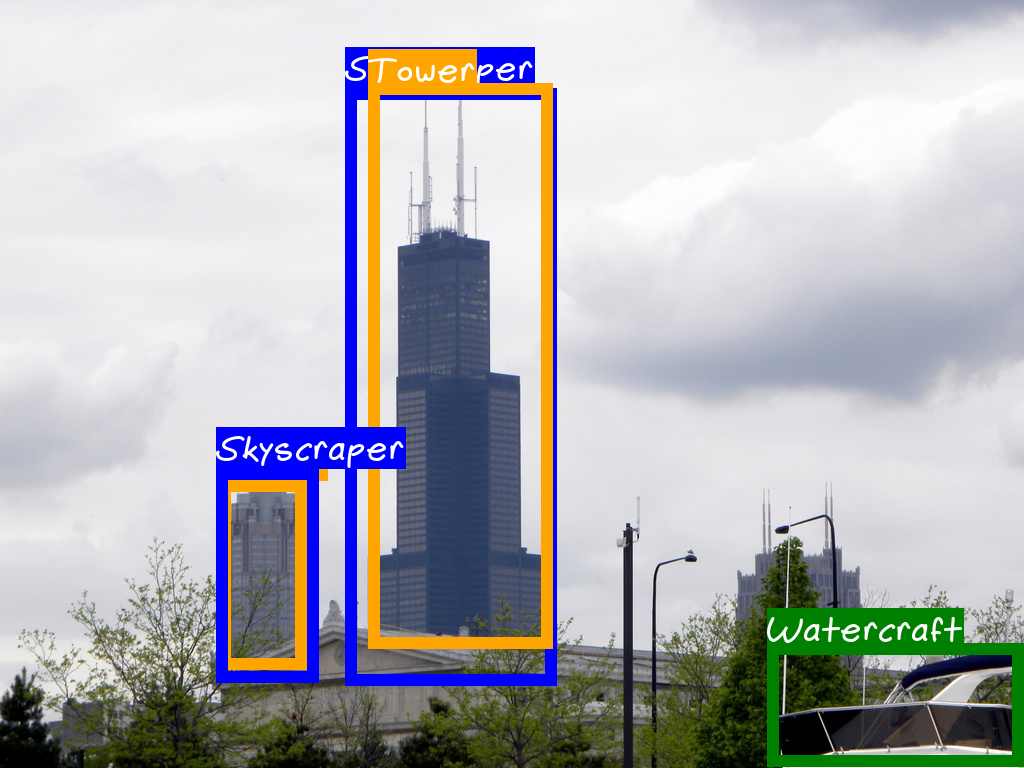}
        \label{fig:overlap_1_1}
    }
    \subfloat[After]{
        \includegraphics[width=0.45\linewidth]{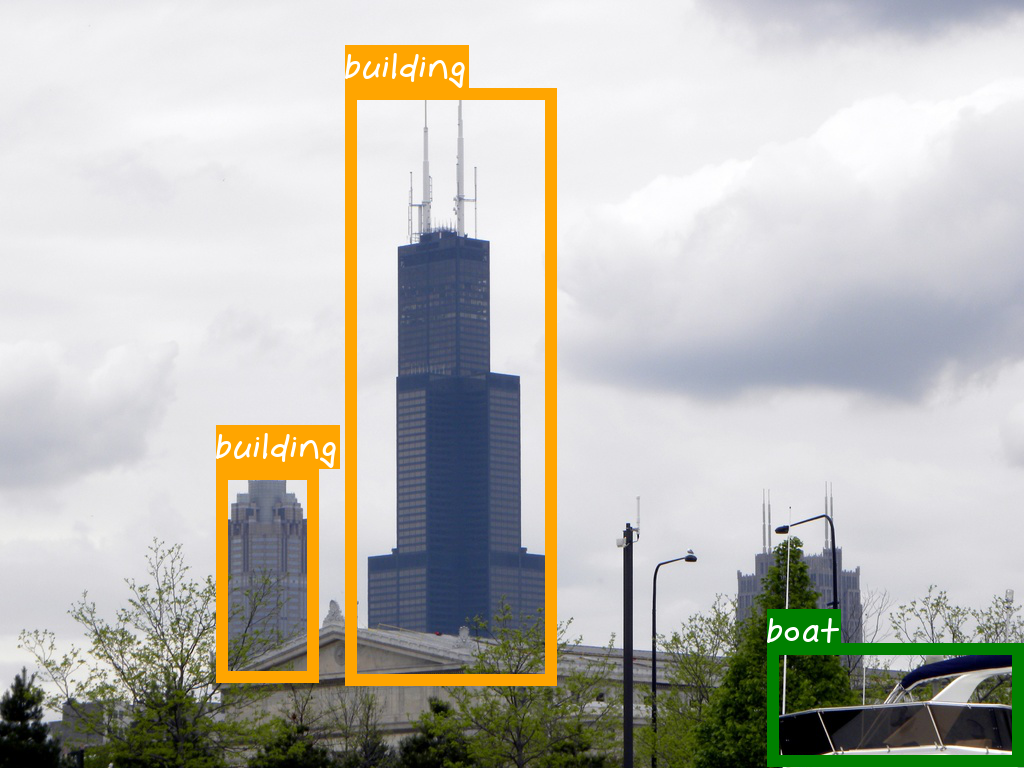}
        \label{fig:overlap_1_2}
    }
    
    \subfloat[Before]{
        \includegraphics[width=0.45\linewidth]{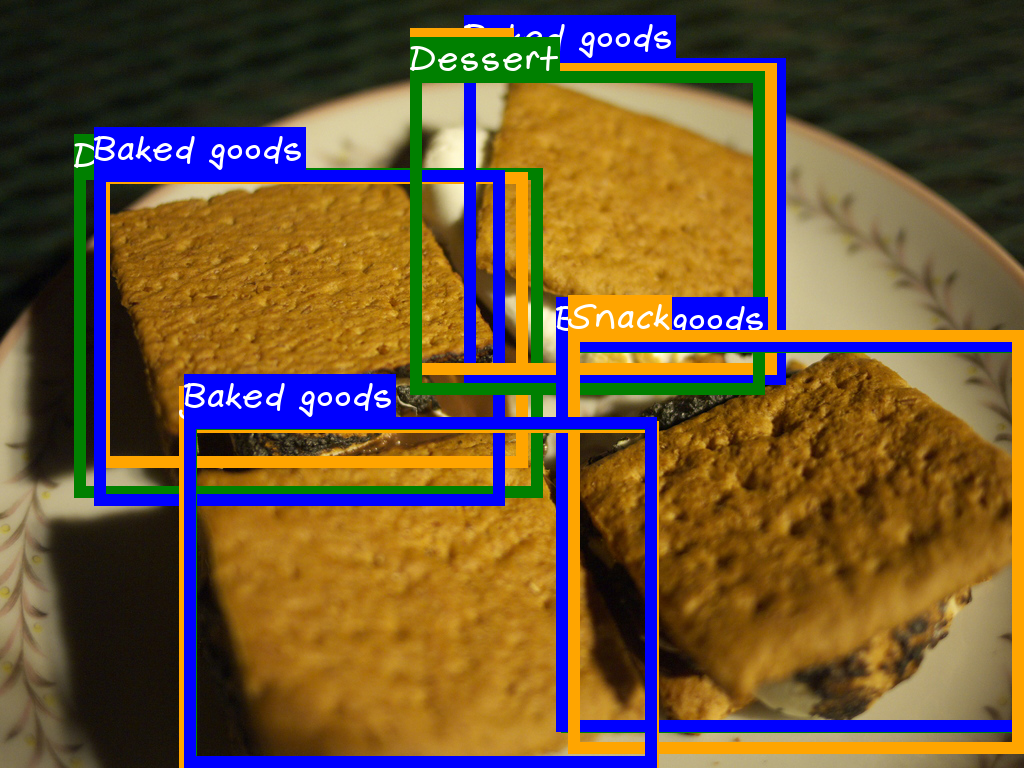}
        \label{fig:overlap_2_1}
    }
    \subfloat[After]{
        \includegraphics[width=0.45\linewidth]{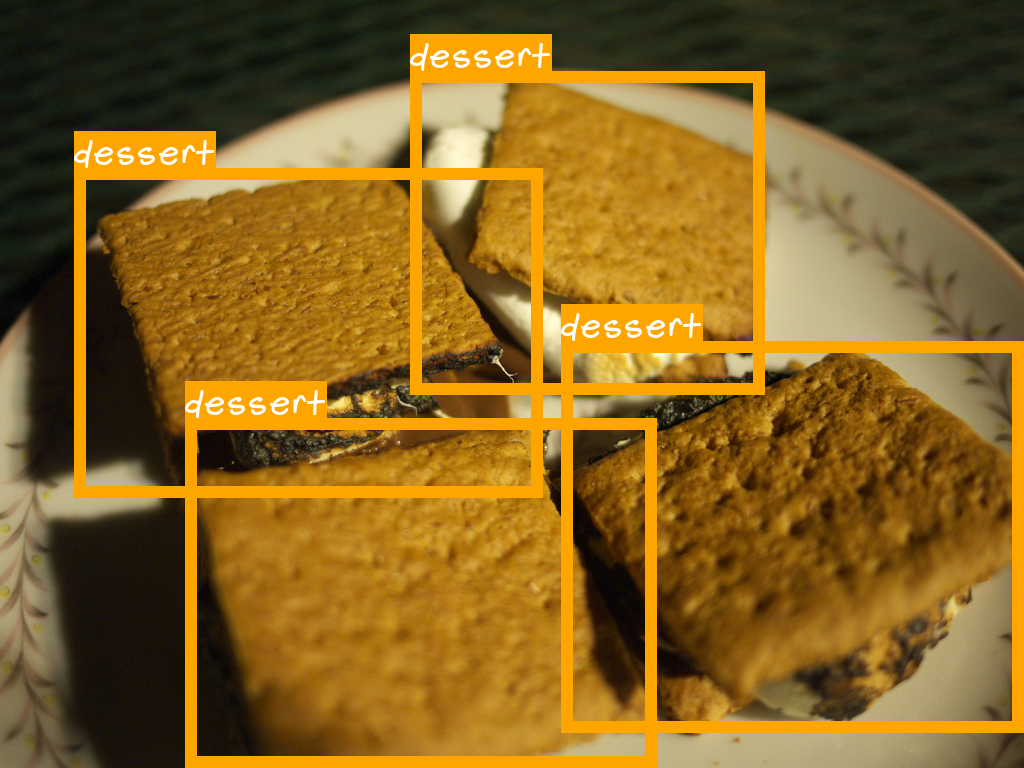}
        \label{fig:overlap_2_2}
    }
    
    \caption{Visual examples of bounding box overlapping problem in OpenImages dataset. Left column: original annotations with multiple boxes over the same object. Right column: annotations after our bounding boxes de-overlapping.}
    \label{fig:bbox_overlap}
\end{figure}
% \noindent \textbf{Object detectors}
\paragraph{Object detectors}
Given these well annotated datasets, deep learning based object detectors have made significant progress over the past decade.
Based on the network design, existing object detectors using convolutional neural networks (CNNs) can be roughly divided into two types: single-stage detector \cite{liu2016ssd,fu2017dssd,li2017fssd, redmon2016you, redmon2017yolo9000, farhadi2018yolov3, bochkovskiy2020yolov4, tian2019fcos, law2018cornernet, zhou2019objects} and two-stage detector \cite{girshick2014rich,girshick2015fast,ren2015faster, cai2018cascade, zhou2021probabilistic}. The two-stage models usually offer better performance while the one-stage models run with faster speed.
Recently, with the rise of transformer~\cite{vaswani2017attention, dosovitskiy2020image} in computer vision, some works investigate the combination of CNNs and transformer for improved object detection~\cite{carion2020end,zhu2020deformable,dai2021up,beal2020toward,sun2021rethinking,zheng2020end,wang_2021_pct}
\vspace{-1ex}

\begin{figure}[t]
    \centering
    \subfloat[Before]{
        \includegraphics[width=0.45\linewidth]{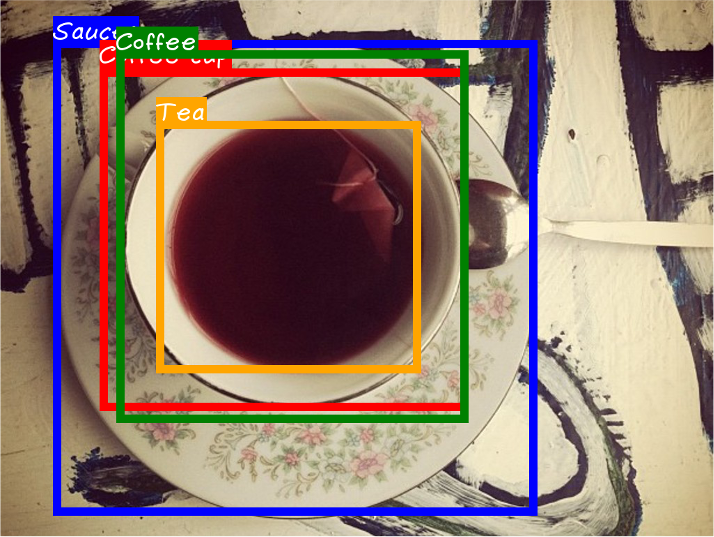}
        \label{fig:invalid_1_1}
    }
    \subfloat[After]{
        \includegraphics[width=0.45\linewidth]{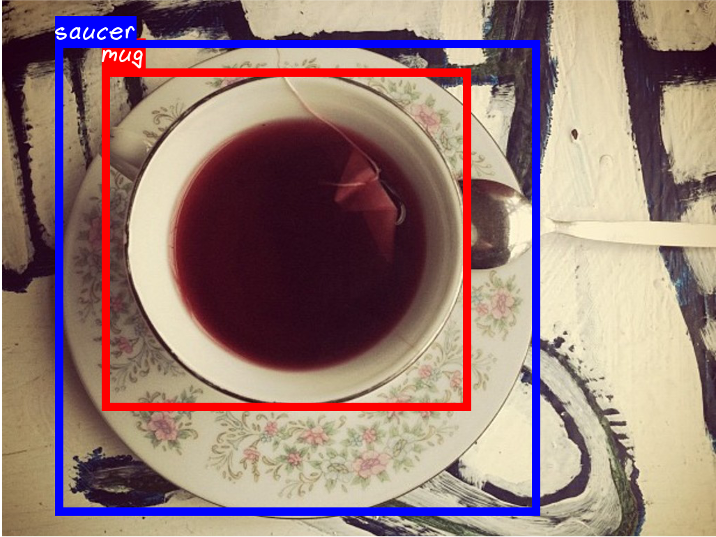}
        \label{fig:invalid_1_2}
    }
    
    \subfloat[Before]{
        \includegraphics[width=0.45\linewidth]{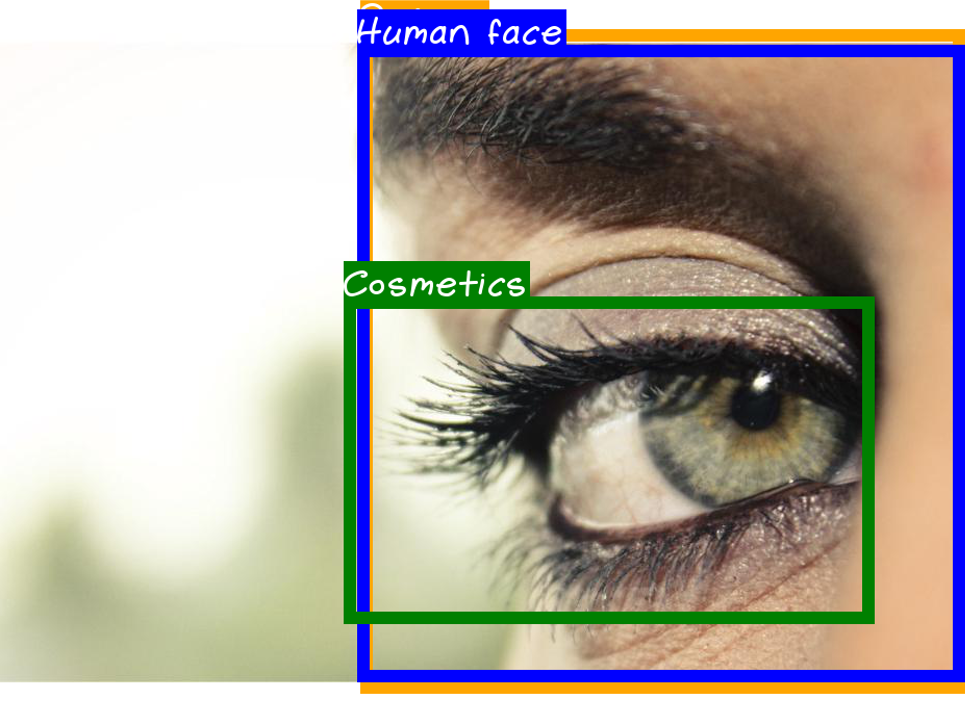}
        \label{fig:invalid_2_1}
    }
    \subfloat[After]{
        \includegraphics[width=0.45\linewidth]{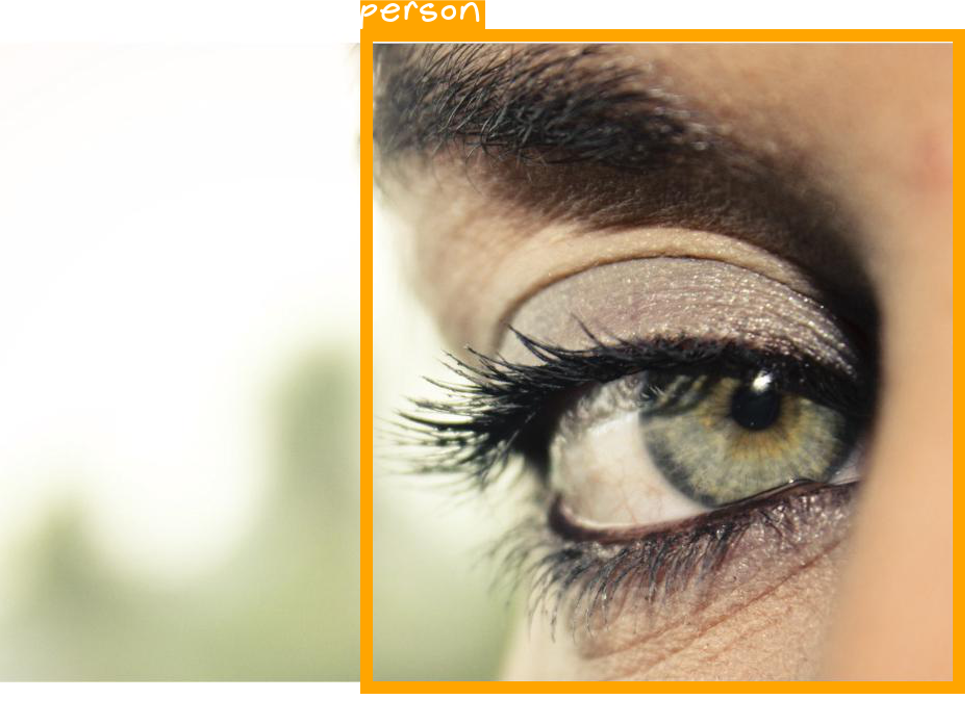}
        \label{fig:invalid_2_2}
    }
    
    \caption{Visual examples of  object categories that might confuse model training, such as ``\textit{Coffee}'' and ``\textit{Coffee cup}'', ``\textit{Cosmetic}'' and ``\textit{Human eye}''. Left column: original annotations. Right column: categories we keep in BigDetection.}
    \label{fig:invalid}
\end{figure}

\section{BigDetection Dataset}
\label{sec:bigdet}
\begin{figure*}[t]
    \centering
    \subfloat[]{
        \includegraphics[width=0.48\linewidth]{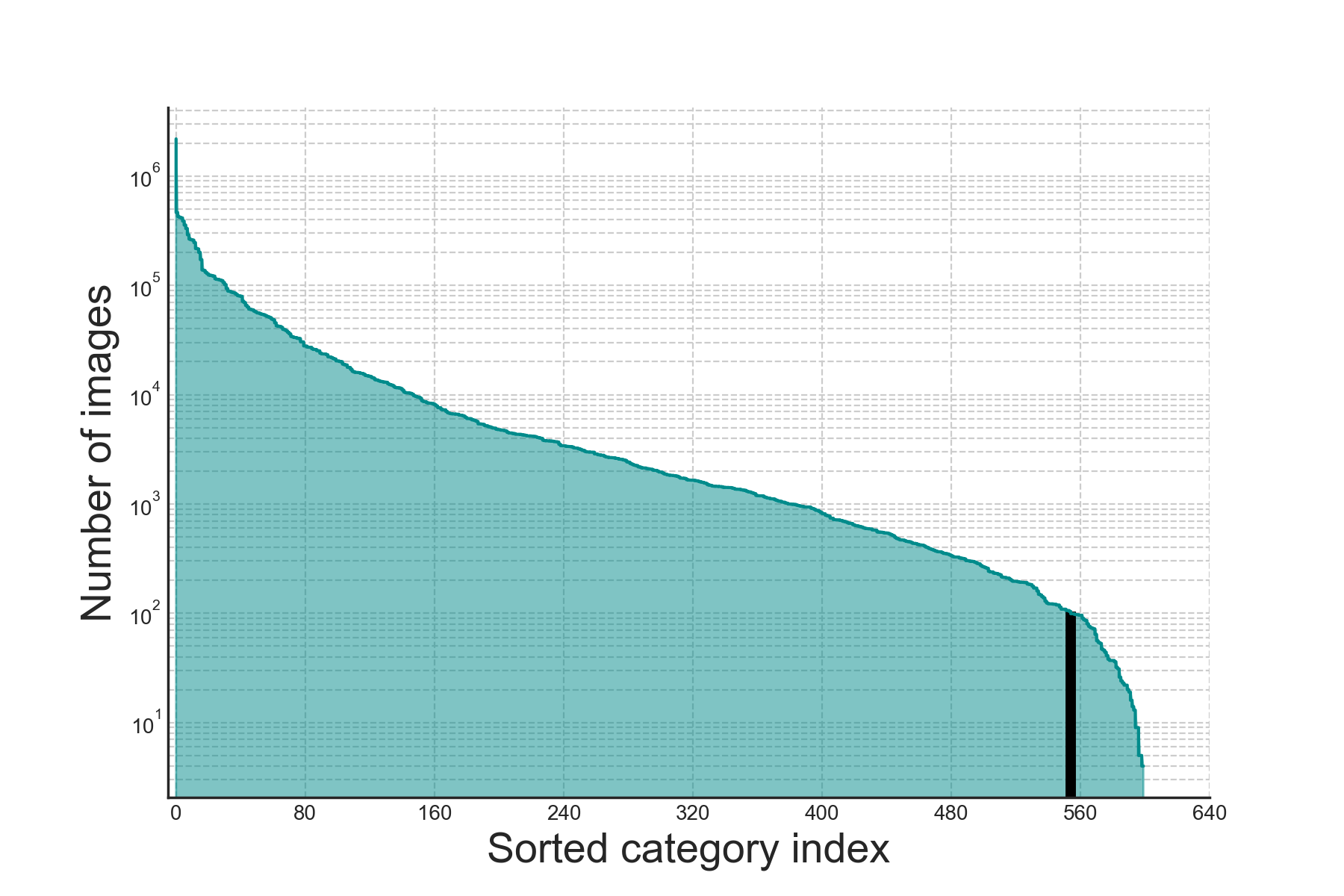}
        \label{fig:stat_1}
    }
    \subfloat[]{
        \includegraphics[width=0.48\linewidth]{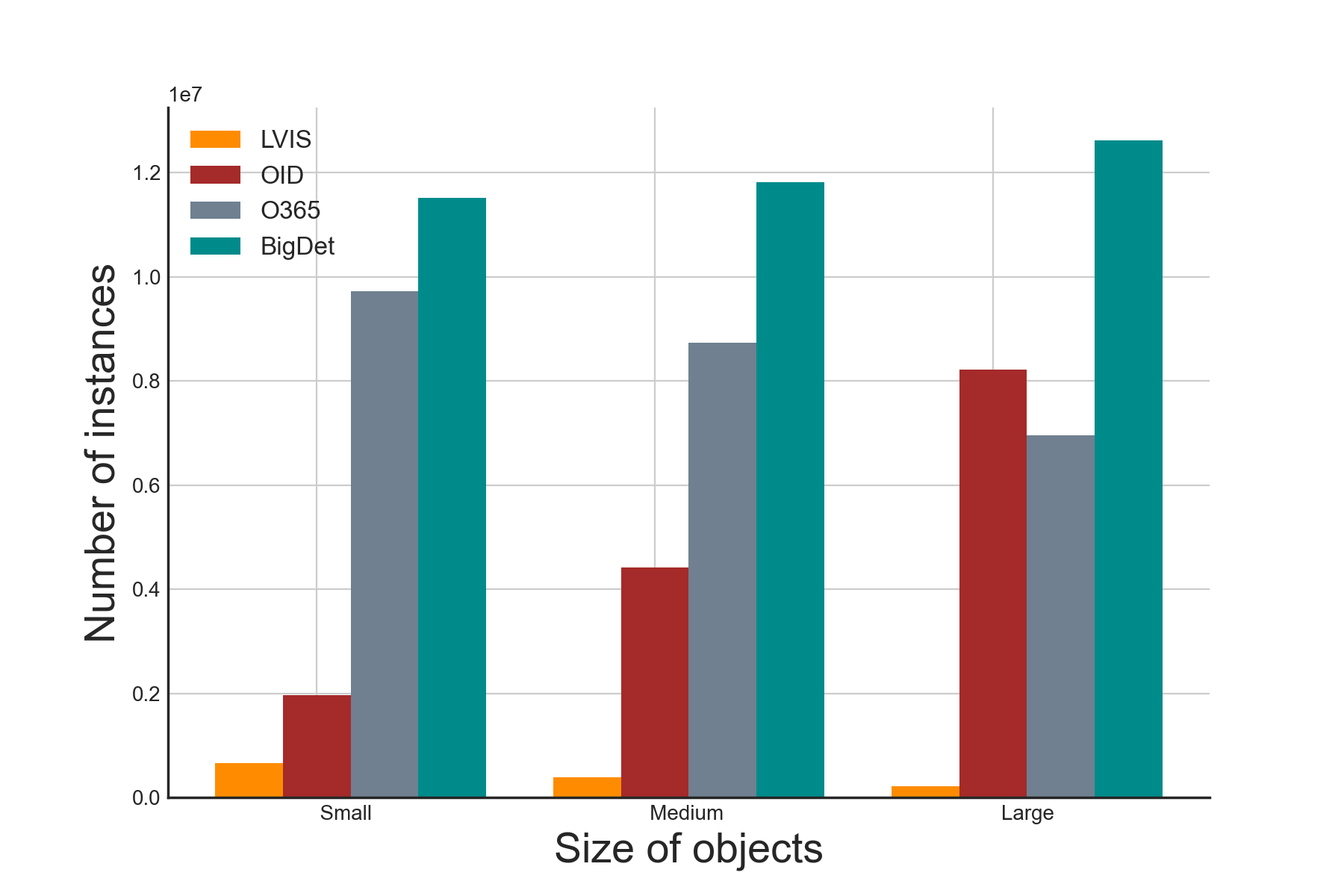}
        \label{fig:stat_2}
    }
    \vspace{-2ex}
    \caption{ (a) Number of images per category of BigDetection. BigDetection have $555$ frequent categories (black line) out of $600$, which means it suffers less from long-tail problem. (b) Number of instances in different object sizes. We find that OpenImages and Objects365 are biased towards certain scale, while BigDetection is balanced across object scales.}
    \label{fig:statistics}
    \vspace{-2ex}
\end{figure*}

The goal of this work is to construct an evolving object detection benchmark designed to incubate next generation object detectors. Our basic idea is to simply leverage the training data from several existing datasets, with carefully designed principles to construct a larger dataset more suitable for pre-training. 

\subsection{Existing Datasets and Limitations}
\label{subsec:data_and_limit}
We first give a brief review on three existing large-scale object detection datasets LVIS~\cite{gupta2019lvis},  OpenImages~\cite{kuznetsova2020open} and Objects365~\cite{shao2019objects365}. All three datasets have been widely used for object detection pre-training. 

\vspace{1ex}
\noindent \textbf{LVIS V1.0} 
LVIS is a dataset designed for large vocabulary instance recognition. It collects high-quality object bounding boxes and segmentation masks for over $1200$ object categories using samples of COCO~\cite{lin2014microsoft}.
However, LVIS naturally has an extremely long-tailed distribution. Nearly half of the categories in LVIS have few training examples (e.g., $\le 20$). 
Besides, given its object categories are more than 10 times of COCO, LVIS has some uninformative annotations, such as the ``\textit{crumb}'' example in Fig~\ref{fig:lvis_oid_badcase}. 
Both attributes make LVIS unsuitable as a pre-training dataset.

\noindent \textbf{OpenImages V6} 
OpenImages (OID) is a large-scale dataset of about 9M images with rich annotations, including image-level labels, object bounding boxes, object segmentation masks, visual relationships, localized narratives, etc.
In terms of object detection, OpenImages has $14.6$M bounding boxes over $600$ object classes.
$90\%$ of these boxes are manually drawn by professional annotators using clicking interface~\cite{papadopoulos2017extreme}, while the remaining $10\%$ are produced semi-automatically using~\cite{papadopoulos2016we}.
We find that there still exists a fair amount of annotations with poor quality.
For example, we observe bounding box overlapping problem. As can be seen in \cref{fig:overlap_1_1} and \cref{fig:overlap_2_1}, several bounding boxes with similar size locate at the same place, but have different class labels. This may confuse model training. 
We also argue that some categories in OpenImages may not be useful for general detector pre-training, such as ``\textit{tea}'' and ``\textit{cosmetics}'' in \cref{fig:invalid_1_1} and \cref{fig:invalid_2_1}.

\noindent \textbf{Objects365} 
Objects365 is another large-scale dataset designed for object detection pre-training. 
It contains around $1.72$M images with more than $22.8$M bounding boxes over 365 categories.
Comparing with OpenImages, Objects365 is close in terms of dataset scale, but has a smaller vocabulary. 
This may not cover enough semantic concepts to pre-train a universal object detector and generalize to other domains.

\subsection{Building a Unified Label Space}
\label{subsec:build}

Despite being large-scale, these three datasets have their own limitations and challenges, such as LVIS being too fine-grained, noisy annotations in OpenImages and relatively small number of object categories for Objects365. 
It would be ideal if we can find a way to combine the datasets and alleviate their individual limitations.
However, this is non-trivial given the heterogeneous label spaces.

As we mentioned in Sec.~\ref{sec:related}, there are some studies on multi-dataset detector training, such as using split classifiers~\cite{wang2019towards,zhou2021simple} and pseudo labeling~\cite{zhao2020object}. But considering the noisy annotations and domain gap among different datasets, we would like to clean the data before model training and combine datasets in a more careful manner.

Our goal is to merge the datasets under one unified label space, and train a single detector on it. In order to create the unified label space, we introduce a category mapping pipeline using language models, which is illustrated in \cref{fig:cat_mapping}.
First, we adopt LVIS's object categories as the initial vocabulary, since LVIS dataset has the largest taxonomy with the most fine-grained annotations.
Second, we utilize a Bert-Large model\footnote{\url{https://huggingface.co/bert-large-uncased}}~\cite{devlin2018bert} to extract features of category words in each dataset.
Third, we compute a cosine similarity between each category word of Objects365/OpenImages and that of LVIS. 
The intuition is the more similar the features are, the higher possibility those categories can be merged. 
Thus, an initial category mapping dictionary will be generated by collecting the top-$10$ similar pairs. 
In the end, to further enhance the validity of the final vocabulary, we manually verify each matching pair in the dictionary with the following principles:

\noindent \textbf{Classes matching} 
We notice that some object categories should belong to the same semantic concept, but their feature similarity is low due to different category words, such as ``\textit{Remote}'' (Objects365) and ``\textit{remote\_control}'' (LVIS). 
In this case, we will perform a manual merge.
For some categories in OpenImages and Objects365 that never occurred in LVIS, we will just adopt them as new classes.

\noindent \textbf{Classes merging}
Since the class granularity of each dataset is different, some categories have inclusion relationship in semantic space. In order to obtain a unified label space, we simply merge these categories into a single one. For example, we merge different bird specifies to the  ``\textit{bird}'' class.

\noindent \textbf{Classes removing} 
We argue that some non-discriminating categories or classes with too few training examples are not suitable for general detector pre-training. 
These classes will be directly removed. 
Some examples are illustrated in~\cref{fig:lvis_oid_badcase}.

\noindent \textbf{BBoxes de-overlapping} 
We find that even after removing some classes, there still exists a large number of overlapped bounding boxes. 
In order to filter them credibly, we first collect all category pairs with bounding boxes IoU greater than $0.65$. 
Then for object categories that always co-occur, we keep them if they are supposed to be multi-labels for the same object. Otherwise we remove them from the annotation set.
We show some de-overlapping results in \cref{fig:overlap_1_2} and \cref{fig:overlap_2_2}. 

\subsection{BigDetection and Its Statistics}
\label{subsec:statistics}

At this point, we have a unified label space that can combine the training data of Objects365, OpenImages, and LVIS.
This allows us to build the largest publicly available object detection dataset, thus we name it BigDetection.
% largest existing object detection dataset, thus we name it BigDetection. 

In \cref{tab:dataset}, we show its statistics comparison to several other large-scale datasets. 
In terms of training set, BigDetection has around 36M bounding boxes in 3.4M images for 600 object categories. On average, there are 10.3 annotated bounding boxes per image.

In addition, we plot the number of images in each category in~\cref{fig:stat_1}.
According to LVIS~\cite{gupta2019lvis}, a category is considered \textit{frequent} if it contains more than $100$ samples.
% considered as \textit{frequent} if there are more than $100$ images it appears.
In BigDetection, we have $555$ \textit{frequent} classes, which surpasses OpenImages ($540$) and Objects365 ($363$). 
Since the majority of classes are frequent, BigDetection suffers less from long-tail problem, which makes it more ideal for object detector pre-training.

In terms of object sizes, we plot the number of instances in each scale bin\footnote{The three object scales follow the definition in COCO~\cite{lin2014microsoft}: $\text{Small} < 32\times32$, $32\times32 < \text{Medium} < 96\times96$, $96\times96 < \text{Large}$} for different datasets in \cref{fig:stat_2}.
We can see that OpenImages and Objects365 are biased towards certain scale, while BigDetection is balanced across object scales.
We will show later that this property helps detector in reducing localization errors.

\section{Pre-training on BigDetection}
\label{sec:analysis}
\begin{table*}[t]
    \begin{center}
        \begin{tabular}{c|c|ccc|cc}
            \toprule
            \multirow{2}*{Detector} & \multirow{2}*{Backbone} & \multicolumn{3}{c|}{BigDetection} & \multicolumn{2}{c}{COCO} \\
            ~ & ~ & AP & AP$_{50}$ & AP$_{75}$ & AP & AP$^*$ \\
            \midrule
            YOLOv3~\cite{farhadi2018yolov3} & D53 & 9.7 & 17.4 & 9.7 & 21.8 & 30.5({\color{red} +8.7}) \\
            Deformable DETR & R50 & 13.1 & 19.3 & 14.2 & 37.4 & 39.9({\color{red} +2.5}) \\
            Faster R-CNN~\cite{ren2015faster} & R50 & 18.9 & 28.8 & 20.5 & 35.7 & 38.8({\color{red} +3.1}) \\
            Faster R-CNN~\cite{ren2015faster} & R50-FPN  & 19.4 & 29.3 & 21.3 & 37.9 & 40.5({\color{red} +2.6}) \\
            CenterNet2~\cite{zhou2021probabilistic} & R50-FPN    & 23.1 & 30.2 & 24.9 & 42.9 & 45.3({\color{red} +2.4}) \\
            Cascade R-CNN~\cite{cai2018cascade} & R50-FPN & 24.1 & 33.0 & 25.8 & 42.1 & 45.1({\color{red} +3.0}) \\
            \bottomrule
        \end{tabular}
    \end{center}
    \caption{BigDetection as a challenging and effective pre-training new benchmark. First, we provide comparison of popular object detection methods on BigDetection validation. All models are trained with an 8$\times$ schedule to enable fair comparison. Then we show the finetuning results on COCO validation set after 1$\times$ finetuning. AP$^*$ indicates that models are pre-trained on BigDetection.}
    \label{tab:cnn}
\end{table*}

Large datasets are usually good for model pre-training, but they also pose challenges.
For example, BigDetection has a serious class imbalance problem. Some classes like ``\textit{person}'' have greatly more annotations than others like ``\textit{ferret}''.
In addition, BigDetection suffers from the partial annotation problem when merging the datasets. Some object categories annotated in one dataset could be considered as ``\textit{background}'' in another dataset. 
In this section, we investigate effective methods to handle class imbalance and partial annotation problems during model training.

\subsection{Class Imbalance}
\label{subsec:imbalance}
There are several widely adopted strategies to alleviate class imbalance problem, like loss re-weighting~\cite{tan2020equalization, li2020overcoming, wang2021seesaw}, data re-sampling~\cite{chang2021image, shen2016relay, gupta2019lvis} and data augmentation~\cite{ghiasi2021simple, zhang2021mosaicos}. In this work, we explore all of them and find that data re-sampling, especially class-aware sampling~\cite{shen2016relay}, is most effective.

To be specific, we use fixed class weights derived from the dataset to perform loss re-weighting. The more samples one class contains, the lower weight will be assigned to that class when computing the loss. However, this does not help the training since BigDetection is so imbalanced, which leads to slow convergence. For data augmentation, we adopt the recent CopyPaste~\cite{ghiasi2021simple} augmentation who achieves great performance on instance segmentation. 
The core idea of CopyPaste is to randomly paste masked objects from one image onto another inside a training batch. 
Unfortunately, BigDetection only has bounding box annotations. Directly pasting the boxed image patches will inevitably introduce unnecessary noise. 
For data re-sampling, we use class-aware sampling (CAS) method following~\cite{shen2016relay}. CAS samples each class with equal probability, which is ideal for datasets with imbalanced classes. Note that since each sample contains multiple instances of different categories, the sampled data will not be completely balanced. 
We will present the experimental results in later Sec.~\ref{subsec:ablation}.

\subsection{Partial Annotations}
\label{subsec:self-training}
In order to address the partial annotation problem when merging different datasets, we adopt a self-training approach similar to ~\cite{zhao2020object,zhu_2021_csst} to complement the ground truth annotations. The goal is to generate additional pseudo annotations that were not manually labeled in the original images. 

In our work, using self-training is more straightforward than~\cite{zhao2020object} since we already have a unified label space.
To be specific, we first train a teacher model on BigDetection.
Then the teacher model is used to generate pseudo annotations for the train set of BigDetection.
Noted for object detection, pseudo annotations include two elements: pseudo labels for classification, and pseudo bounding boxes for localization.
The credibility of pseudo labels and the maximum overlap area of pseudo boxes can be adjusted by changing the values of score threshold and NMS threshold of the teacher model, respectively.
The last step is to incorporate these new pseudo annotations to the ground truth and train a better student model.
However, in order to ensure that these pseudo boxes capture the missing objects without introducing more noise, we add an additional step to filter the pseudo annotations.
Namely, one pseudo annotation will be removed if the IoU between its box and any ground truth box is greater than $0.6$. 
Once filtered, the remaining pseudo annotations will be used to augment the training set. 

We find that even two detectors have similar mAP on a dataset, their precision for each class differs greatly due to the different training setting. 
To further improve the credibility of pseudo annotations, we adopt a multi-teacher strategy. 
Suppose the ground truth annotation set for sample $i$ is $Y^i_{gt}$, and we have multiple teacher models $[t_1, t_2, \cdots, t_k]$. Each teacher model generates a pseudo annotation set $Y^i_{t_j}, j=1,\cdots,k$. 
The final annotation set of sample $i$ will be obtained:
\begin{equation*}
    \widetilde{Y}^i = \text{NMS}(Y^i_{t_1},\cdots, Y^i_{t_k}; \tau)\cup Y^i_{gt}
\end{equation*}
where NMS is utilized to de-overlap multiple pseudo annotation sets and $\tau$ is the threshold. We set $k=2$ throughout this work. More details can be found in the \ref{app:pseudo_anno}.
\begin{figure}[t]
  \centering
  \includegraphics[width=.48\linewidth]{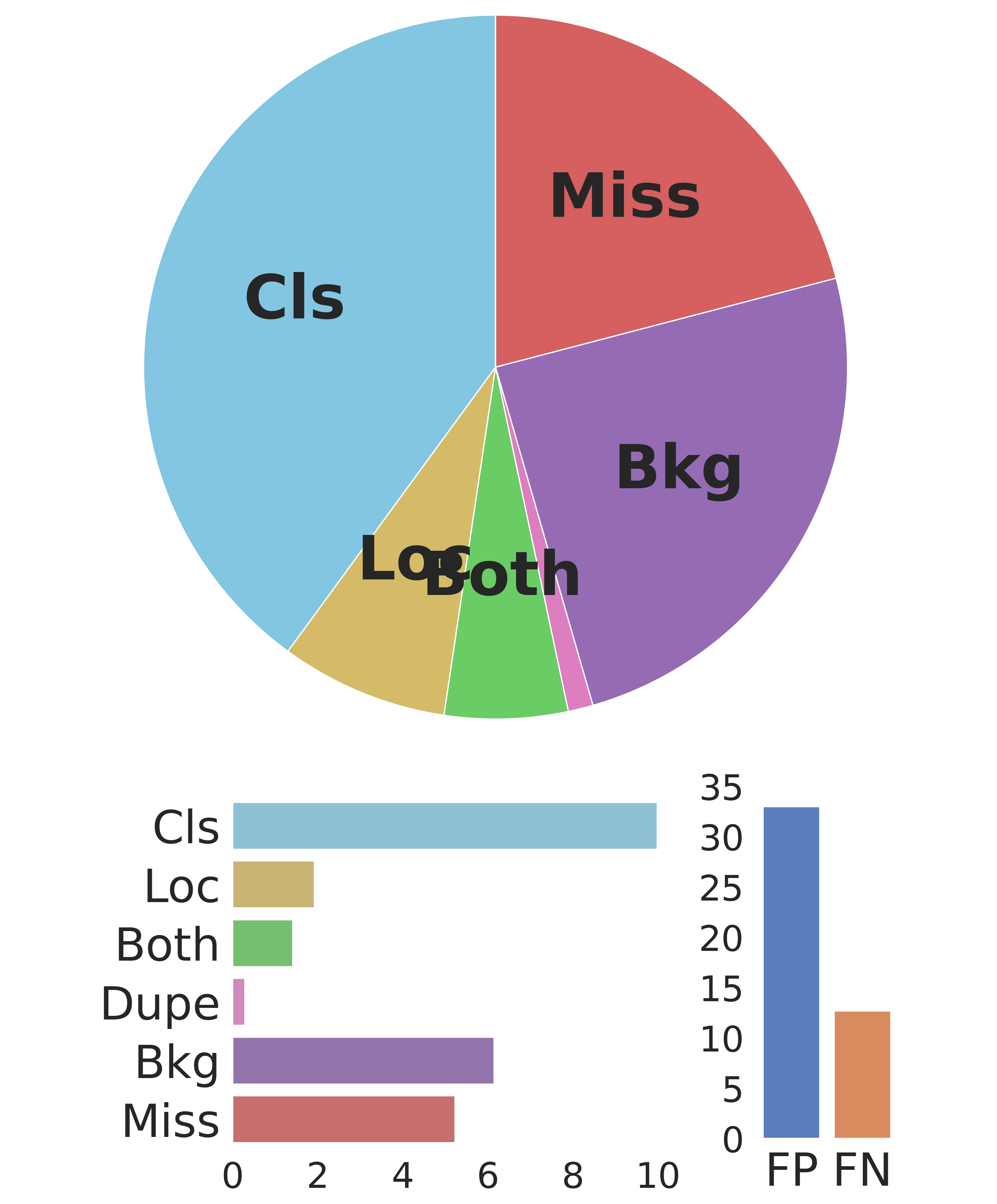}
  \includegraphics[width=.48\linewidth]{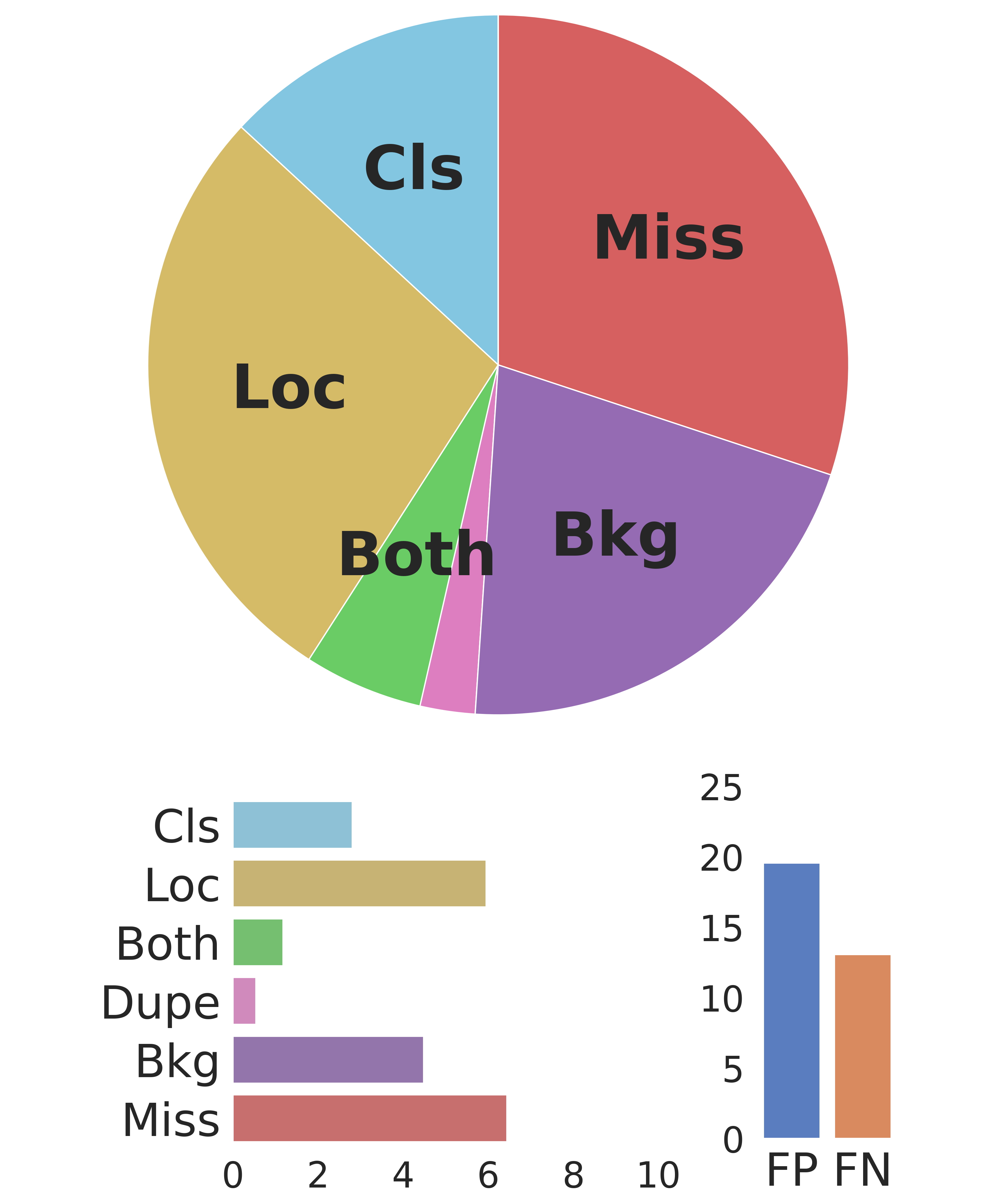}
  \caption{Error diagnose by TIDE~\cite{tide-eccv2020}. Left: BigDetection. Right: COCO.}
  \label{fig:error_ana}
  \vspace{-3ex}
\end{figure}

\section{Experiments}
\label{sec:experiments}

\begin{table*}[t]
    \begin{center}
        \begin{tabular}{lcc|ccccccc}
            \toprule
            Method & Backbone & TTA & AP$_{\text{val}}$ & AP$_{\text{test}}$ & AP$_{50}$ & AP$_{75}$ & AP$_S$ & AP$_M$ & AP$_L$   \\
            \midrule
            SpineNet~\cite{du2020spinenet} & SpineNet-190 & \ding{55} & 52.2 & 52.5 & - & - & - & - & - \\
            RelationNet++~\cite{chi2020relationnet++} & ResNeXt-101-DCN & \checkmark & - & 52.7 & 70.4 & 58.3 & 35.8 & 55.3 & 64.7 \\
            ResNeSt~\cite{zhang2020resnest} & ResNeSt-200 & \checkmark & 52.5 & 53.3 & 72.0 & 58.0 & 35.1 & 56.2 & 66.8 \\
            CBNet~\cite{liu2020cbnet} & ResNeXt152 & \checkmark & - & 53.3 & 71.9 & 58.5 & 35.5 & 55.8 & 66.7 \\ 
            EfficientDet-D7x~\cite{tan2020efficientdet} & EfficientNet-D7x & \ding{55} & 54.4 & 55.1 & 74.3 & 59.9 & 37.2 & 57.9 & 68.0 \\
            DetectoRS~\cite{qiao2021detectors} & ResNeXt-101-DCN & \checkmark & - & 55.7 & 74.2 & 61.1 & 37.7 & 58.4 & 68.1 \\
            ScaledYOLOv4~\cite{wang2021scaled} & CSP-P7 & \checkmark & - & 55.8 & 73.2 & 61.2 & - & - & - \\
            CenterNet2$\dagger$~\cite{zhou2021probabilistic} & Res2Net-101-DCN & \ding{55} & 56.1 & 56.4 & 74.0 & 61.6 & 38.7 & 59.7 & 68.6 \\
            CopyPaste$\dagger$~\cite{ghiasi2021simple} & Efficient-B7 & \ding{55} & 57.0 & 57.3 & - & - & - & - & - \\ 
            HTC++~\cite{liu2021swin} & Swin-B & \ding{55} & 56.4 & - & - & - & - & - & - \\
            CBNetV2~\cite{liang2021cbnetv2} & Swin-B & \ding{55} & 58.4 & 58.7 & 76.9 & 64.3 & 40.7 & 62.0 & 72.0 \\
            CBNetV2~\cite{liang2021cbnetv2} & Swin-B & \checkmark & 58.9 & 59.3 & 77.6 & 65.0 & 41.7 & 62.7 & 72.5  \\
            \textbf{CBNetV2 (BigDet)} & Swin-B & \ding{55} & 59.1 & 59.5 & 77.3 & 65.3 & 42.0 & 62.4 & 72.7 \\
            \textbf{CBNetV2 (BigDet)} & Swin-B & \checkmark & \textbf{59.5} & \textbf{59.8} & \textbf{77.9} & \textbf{65.6} & \textbf{42.2} & \textbf{62.9} & \textbf{73.0} \\
            \midrule
            \textcolor{gray}{HTC++}~\cite{liu2021swin} & \textcolor{gray}{Swin-L} & \textcolor{gray}{\ding{55}} & \textcolor{gray}{57.1} & \textcolor{gray}{57.7} & \textcolor{gray}{-} & \textcolor{gray}{-} & \textcolor{gray}{-} & \textcolor{gray}{-} & \textcolor{gray}{-} \\
            \textcolor{gray}{HTC++}~\cite{liu2021swin} & \textcolor{gray}{Swin-L} & \textcolor{gray}{\checkmark} & \textcolor{gray}{58.0} & \textcolor{gray}{58.7} & \textcolor{gray}{-} & \textcolor{gray}{-} & \textcolor{gray}{-} & \textcolor{gray}{-} & \textcolor{gray}{-} \\
            \textcolor{gray}{DyHead}~\cite{yang2021focal} & \textcolor{gray}{Focal-L} & \textcolor{gray}{\checkmark} & \textcolor{gray}{58.7} & \textcolor{gray}{58.9} & \textcolor{gray}{-} & \textcolor{gray}{-} & \textcolor{gray}{-} & \textcolor{gray}{-} & \textcolor{gray}{-} \\
            \textcolor{gray}{CBNetV2}~\cite{liang2021cbnetv2} & \textcolor{gray}{Swin-L} & \textcolor{gray}{\ding{55}} & \textcolor{gray}{59.1} & \textcolor{gray}{59.4} & \textcolor{gray}{-} & \textcolor{gray}{-} & \textcolor{gray}{-} & \textcolor{gray}{-} & \textcolor{gray}{-} \\
            \textcolor{gray}{CBNetV2}~\cite{liang2021cbnetv2} & \textcolor{gray}{Swin-L} & \textcolor{gray}{\checkmark} & \textcolor{gray}{59.6} & \textcolor{gray}{60.1} & \textcolor{gray}{-} & \textcolor{gray}{-} & \textcolor{gray}{-} & \textcolor{gray}{-} & \textcolor{gray}{-} \\
            \textcolor{gray}{DyHead}~\cite{dai2021dynamic} & \textcolor{gray}{Swin-L} & \textcolor{gray}{\checkmark} & \textcolor{gray}{58.4} & \textcolor{gray}{58.7} & \textcolor{gray}{77.1} & \textcolor{gray}{64.5} & \textcolor{gray}{41.7} & \textcolor{gray}{62.0} & \textcolor{gray}{72.8} \\
            \textcolor{gray}{DyHead}$\dagger$~\cite{dai2021dynamic} & \textcolor{gray}{Swin-L} & \textcolor{gray}{\checkmark} & \textcolor{gray}{60.3} & \textcolor{gray}{60.6} & \textcolor{gray}{78.5} & \textcolor{gray}{66.6} & \textcolor{gray}{43.9} & \textcolor{gray}{64.0} & \textcolor{gray}{74.2} \\
            \textcolor{gray}{Soft Teacher}~\cite{xu2021end} & \textcolor{gray}{Swin-L} & \textcolor{gray}{\ding{55}} & \textcolor{gray}{59.1} & \textcolor{gray}{-} & \textcolor{gray}{-} & \textcolor{gray}{-} & \textcolor{gray}{-} & \textcolor{gray}{-} & \textcolor{gray}{-} \\
            \textcolor{gray}{Soft Teacher}$\dagger$~\cite{xu2021end} & \textcolor{gray}{Swin-L} & \textcolor{gray}{\checkmark} & \textcolor{gray}{60.7} & \textcolor{gray}{61.3} & \textcolor{gray}{-} & \textcolor{gray}{-} & \textcolor{gray}{-} & \textcolor{gray}{-} & \textcolor{gray}{-} \\
            \bottomrule
        \end{tabular}
    \end{center}
    \vspace{-2ex}
    \caption{Comparison with state-of-the-art object detection methods on COCO validation and test-dev sets. BigDet: pretrained on BigDetection dataset. TTA: test-time augmentation. $\dagger$ indicates method trained with extra unlabeled data. Our CBNetv2 with Swin-B backbone achieves 59.8 AP on COCO test-dev. This result almost reaches a CBNetv2 model with Swin-Large backbone without BigDetection pretraining, and surpasses most methods using Swin-Large backbone (bottom section).}
    \label{tab:swin}
    \vspace{-2ex}
\end{table*}

\noindent \textbf{Setup and evaluation protocol}
BigDetection is split into train set~(\verb+bigdet_train+) and validation set~(\verb+bigdet_val+). 
When using it as a new benchmark, we train different detection models on \verb+bigdet_train+ and evaluate their performance on \verb+bigdet_val+. 
When using it as a pre-training dataset, we first pre-train the detection models on \verb+bigdet_train+, and then finetune them on COCO train set, and report results on either COCO validation or test-dev set. 
For both evaluations on \verb+bigdet_val+ and COCO, we follow the standard COCO style metrics to report mean average precision (mAP) under different IoU thresholds and object scales.
We also adopt a partially labeled data setting to study data efficiency. Pre-trained models will be finetuned on COCO using $1\%$, $2\%$, $5\%$ and $10\%$ labeled data.

\noindent \textbf{Implementation details}
We adopt CenterNet2~\cite{zhou2021probabilistic} equipped with ResNet-50 and FPN~\cite{lin2017feature} to provide baseline results and conduct ablation study. 
Our implementation is based on Detectron2~\cite{wu2019detectron2}.
Most hyperparameters follow the default setting of CenterNet2 unless otherwise stated. 
Specifically, we train the detector with an SGD optimizer for $8\times$ (720K iterations) on BigDetection pre-training and $1\times$ (90K iterations) on COCO finetuning. 
Base learning rate is set to $0.02$ and is dropped at iterations 660K/60K and 700K/80K.
We use 8 V100 GPUs, with 2 samples per GPU. 
Multi-scale training is adopted with the short edge in range $[640, 800]$ and the long edge up to 1333. 
No extra data augmentations are used, such as Jittering, Mosaic~\cite{zhang2021mosaicos}, CopyPaste~\cite{ghiasi2021simple} or Mix-up~\cite{zhang2017mixup}.

When comparing to state-of-the-art detectors on COCO, we adopt CBNetV2~\cite{liang2021cbnetv2} equipped with a Swin-Transformer-Base backbone. 
For a fair comparison, the training strategy and all hyperparameters follow the default setting in~\cite{liang2021cbnetv2} implemented with MMDetection~\cite{chen2019mmdetection}. Again, we do $8\times$ schedule for pre-training stage on BigDetection and $1\times$ for COCO finetuning.

\subsection{A New Object Detection Benchmark}
\label{subsec:detbench}
To provide a rough picture of how challenging BigDetection is, we select several most popular object detectors to evaluate their performance on \verb+bigdet_val+. The methods include Faster R-CNN~\cite{ren2015faster}, Faster R-CNN with FPN~\cite{lin2017feature}, Cascade R-CNN~\cite{cai2018cascade}, YOLOv3~\cite{farhadi2018yolov3}, CenterNet2~\cite{zhou2021probabilistic} and DETR~\cite{carion2020end}, which represent a variety of object detection models (\eg, two-stage/one-stage, anchor-based/anchor-free, CNN-based/transformer-based).
All these models are trained on \verb+bigdet_train+ for $8\times$ schedule to provide fair comparison. Other hyperparameters follow their default setting in MMDetection~\cite{chen2019mmdetection}.

As can be seen in~\cref{tab:cnn}, good detectors on COCO also perform well on BigDetection, e.g., CenterNet2 and Cascade R-CNN are top performers on both datasets. However, we see that even the best result on BigDetection is 24.1 AP, which is close to the initial results when COCO was introduced in 2014. This suggests that BigDetection is a much more challenging dataset than COCO. We hope BigDetection can help advance the development of next-generation object detection algorithms. 
Unless otherwise stated, we use CenterNet2 for most experiments in the following sections. We find that CenterNet2 often show better generalization during fine-tuning.

In addition, we use our trained CenterNet2 model to perform an error diagnosis by TIDE~\cite{tide-eccv2020}. The results can be visualized in~\cref{fig:error_ana}. By comparing the result on BigDetection (left) and the result on COCO (right), we can see that the main difference lies in the Cls and Loc categories. BigDetection shows more Cls errors since it has much more object categories than COCO. At the same time, BigDetection makes far fewer Loc errors than COCO. Even within BigDetection, Loc errors are fewer than Miss and Bkg. We believe this is because BigDetection is more balanced across object scales as mentioned in~\cref{fig:stat_2}, so that it can better handle small and medium objects. This observation is also supported by results later shown in~\cref{tab:swin} that models pre-trained on BigDetection achieve higher AP$_{S}$ on COCO validation. 

\subsection{Generalization to COCO}
\noindent \textbf{Baseline}
We first show that BigDetection pre-training provides significant benefits for different detector architectures (single-stage or two-stage, anchor-based or anchor-free).
Following the model set in \cref{subsec:detbench}, we finetune each model on COCO train split with 1$\times$ schedule, and ImageNet pre-trained checkpoints are adopted for comparison.
After pre-training on BigDetection, most detectors gain $2\sim3$ AP improvement, and especially YOLOv3 even gains 8.7 AP improvement.
These results suggest that BigDetection forms a strong pre-train dataset to provide better feature representation for downstream transfer.

\noindent \textbf{Comparison to state-of-the-art}
We would like to show how far BigDetection can advance performance of current strongest detectors. We adopt CBNetV2 with Swin-transformer backbone as our baseline~\cite{liang2021cbnetv2}. 

The results are shown in~\cref{tab:swin}. We have several observations. 
First, with pre-training on BigDetection, we can further improve this strong baseline by $0.7$ AP ($58.4$ $\rightarrow$ $59.1$). In particular, most improvements come from small objects, i.e., $AP_{S}$ increases from $40.7$ to $42$. 
Second, combined with test-time augmentation, our CBNetV2 model with Swin-Base backbone pre-trained on BigDetection achieves superior performance on both COCO validation and test-dev sets, $59.5$ and $59.8$ AP respectively. We want to point out that this performance with Swin-Base backbone is even competitive to CBNetv2 with Swin-Large backbone without pre-training on BigDetection. Note that Swin-Large is twice heavier than Swin-Base, which supports well that such pre-training is useful.
Furthermore, this result also surpasses most methods using Swin-Large (\cref{tab:swin} bottom section).
Only~\cite{dai2021dynamic} and~\cite{xu2021end} using extra COCO unlabeled data is marginally better. 

\begin{table}[t]
    \begin{center}
        \begin{tabular}{ccccc}
            \toprule
            Method & $1\%$ & $2\%$ & $5\%$ & $10\%$ \\
            \midrule
            Supervised$\dagger$  & 9.8 & 14.3 & 21.2 & 26.2 \\
            STAC$\dagger$~\cite{sohn2020simple} & 14.0 & 18.3 & 24.4 & 28.6 \\
            SoftTeacher$\dagger$~\cite{xu2021end} & 20.5 & 26.5 & 30.7 & 34.0 \\
            \midrule
            Ours        & \textbf{26.1} & \textbf{29.3} & \textbf{31.9} & \textbf{34.1} \\
            \bottomrule
        \end{tabular}
    \end{center}
    \vspace{-2ex}
    \caption{Comparison with different methods under partially labeled COCO. BigDetection pre-training is particularly beneficial when dealing with insufficient training data. $\dagger$ indicates using strong augmentations, such as Cutout.}
    \label{tab:data-efficiency}
    \vspace{-2ex}
\end{table}
\begin{table}[t]
    \begin{center}
        \begin{tabular}{l|cccc}
            \toprule
             & AP & $\text{AP}_{50}$ & $\text{AP}_{75}$ & COCO \\
            \midrule
            IAS & 12.8 & 17.2 & 13.9 & 44.9 \\
            RFS & 20.4 & 26.9 & 21.9 & 44.2 \\
            CAS & \textbf{23.1} & \textbf{30.2} & \textbf{24.9} & \textbf{45.3} \\
            \bottomrule
        \end{tabular}
    \end{center}
    \vspace{-2ex}
    \caption{Ablation study on the effectiveness of different data samplers used to deal with class imbalanced problem. IAS: instance-aware sampling. RFS: repeated factor sampling. CAS: class-aware sampling. First three columns show results on BigDetection validation set, while the last column shows results on COCO.}
    \label{tab:sampler}
    \vspace{-2ex}
\end{table}
\noindent \textbf{Data-efficiency}
One of the great benefits of pre-training on a large-scale dataset is a well-trained model only needs a few target labels to perform considerably well. Here, we show BigDetection pre-training is helpful across a variety of dataset sizes and helps data efficiency.

Following the partially labeled data setting introduced in STAC~\cite{sohn2020simple}, Faster R-CNN~\cite{ren2015faster} with FPN is adopted for fair comparison. The finetuning is done on COCO using $1\%$, $2\%$, $5\%$ and $10\%$ samples of train split. 
\cref{tab:data-efficiency} summarizes the results.
Our method significantly improves the performance upon supervised baseline and STAC. 
Compared to STAC, we obtain \textbf{12.1}, \textbf{11}, \textbf{7.5} and \textbf{5.5} AP gain under difference dataset sizes. 
We also compare our method to a recent work SoftTeacher~\cite{xu2021end}, which is an end-to-end self-training method for object detection.
Interestingly, our method shows a significant performance improvement (\textbf{5.6} AP) on $1\%$ COCO setting, and performs still better when more data is introduced. 
Note that SoftTeacher uses longer training schedule and strong augmentations.
In \cite{zoph2020rethinking}, the results show that self-training works better than pre-training across dataset sizes with a lowest data regime of $20\%$.
However our work suggests that BigDetection pre-training is particularly useful when dealing with extremely insufficient training data ($1\sim 10\%$). 

\begin{table}[t]
    \begin{center}
        \begin{tabular}{lccc}
            \toprule
             & Sampler & Schedule & AP \\
            \hline
            COCO & - & $1\times$ & 42.9 \\
            COCO & - & $8\times$ & 43.8 \\
            LVIS & RFS & $1\times$+$1\times$ & 37.8 \\
            OpenImages & CAS & $8\times$+$1\times$ & 44.0 \\
            Objects365 & CAS & $8\times$+$1\times$ & 45.1 \\
            \hline
            BigDetection & CAS & $8\times$+$1\times$ & 45.3 \\
            BigDetection$\dagger$ & CAS & $8\times$+$1\times$ & \textbf{45.7} \\
            \bottomrule
        \end{tabular}
    \end{center}
    \vspace{-2ex}
    \caption{Ablation study on generalization ability to COCO using different pre-training datasets. $\dagger$ indicates using additional pseudo annotations generated by self-training. We show that BigDetection serves as a better pre-training dataset. }
    \label{tab:data_ablation}
    \vspace{-2ex}
\end{table}
\subsection{Ablation studies}
\label{subsec:ablation}

\noindent \textbf{Regarding data re-sampling}
Recall in \cref{subsec:imbalance}, data re-sampling is the most effective method to alleviate class imbalance problem. 
Here, we ablate the effects of using different samplers in \cref{tab:sampler}. 

IAS selects each training sample with equal probability, thus it is expected to perform poorly on class imbalanced datasets. 
RFS is designed for low-shot long-tailed data in LVIS~\cite{gupta2019lvis}. It first assigns a pre-computed repeat factor for each category. Then the maximum factor of labeled categories will be chosen for each image. 
Since BigDetection does not show long-tail phenomenon, RFS performs mediocre on \verb+bigdet_val+, and even affects the generalization ability to COCO.
CAS offers the best performance as it is designed to handle class imbalance. It is simple, and thus scales well on large-scale dataset. 

\noindent \textbf{Regarding different pre-training datasets}
It is important to show how BigDetection compares to other large-scale datasets when used as pre-training dataset. 
We use finetuning results on COCO to reflect the capability of the pre-trained models.

In terms of baseline, we use models directly trained on COCO with $1\times$ and $8\times$ schedules.
For other datasets, we always pre-train for $8\times$ schedule and finetune on COCO for $1\times$ schedule to enable fair comparison. CAS is adopted as data sampler for OpenImages, Objects365 and BigDetection, except for LVIS. Since RFS is shown to have more reasonable performance on LVIS. 
As shown in~\cref{tab:data_ablation}, our model pre-trained on BigDetection improves over the baseline by a notable margin ($43.8$ $\rightarrow$ $45.7$). It also outperforms models pre-trained on those individual datasets, which suggests its better potential in pre-training. Furthermore, \cref{tab:data_ablation} also shows individual contributions from using CAS and self-training, i.e., CAS brings $1.5$ AP improvement  while self-training brings another $0.4$ AP improvement.

\section{Limitations}
\label{sec:limitations}

Here we list several limitations that remain unsolved in our dataset.
\textbf{Scalability}
Despite our initial category mapping dictionary is automatically generated with the help of large language models, we heavily rely on manual inspection as described in~\cref{subsec:build} to build the final unified label space.
This is more reliable than machine generated annotations, but also sacrifices scalability to some extent. 
\textbf{Data sampling}
We use CAS to handle class imbalance problem, but it introduces a side problem. Within an $8\times$ schedule, the model may not see all the images in the dataset.
This indicates a low utilization of the data.
\textbf{Noisy annotations}
Following our dataset merging principles, some noisy annotations from OpenImages have been removed, but some remain. 
Our work mainly aims at how to build a unified taxonomy to leverage existing datasets. 
And learning from noisy data will be a promising direction.

\section{Conclusion}
\label{sec:conclusion}
In this paper, we have presented BigDetection, an evolving large-scale object detection dataset. 
It is much larger in multiple dimensions (object categories, training images, bounding box annotations) than previous benchmarks. It could serve as a new challenging benchmark for evaluating different object detection methods, since state-of-the-art detectors only achieve around 30 AP on its validation set. We also show its effectiveness when used as a pre-training dataset. After pre-training on it, we achieve superior generalization performance on COCO validation and test-dev sets, as well as strong data-efficiency results. BigDetection presents both opportunities and challenges. We hope it can be used to incubate next-generation object detectors.

%%%%%%%%% REFERENCES
{\small
\bibliographystyle{ieee_fullname}
\bibliography{main}
}

%%%%%%%%% APPENDIX
\clearpage
\appendix

\section{Pseudo Annotation Details}
\label{app:pseudo_anno}
\begin{figure}[ht]
  \centering
  \includegraphics[width=.8\linewidth]{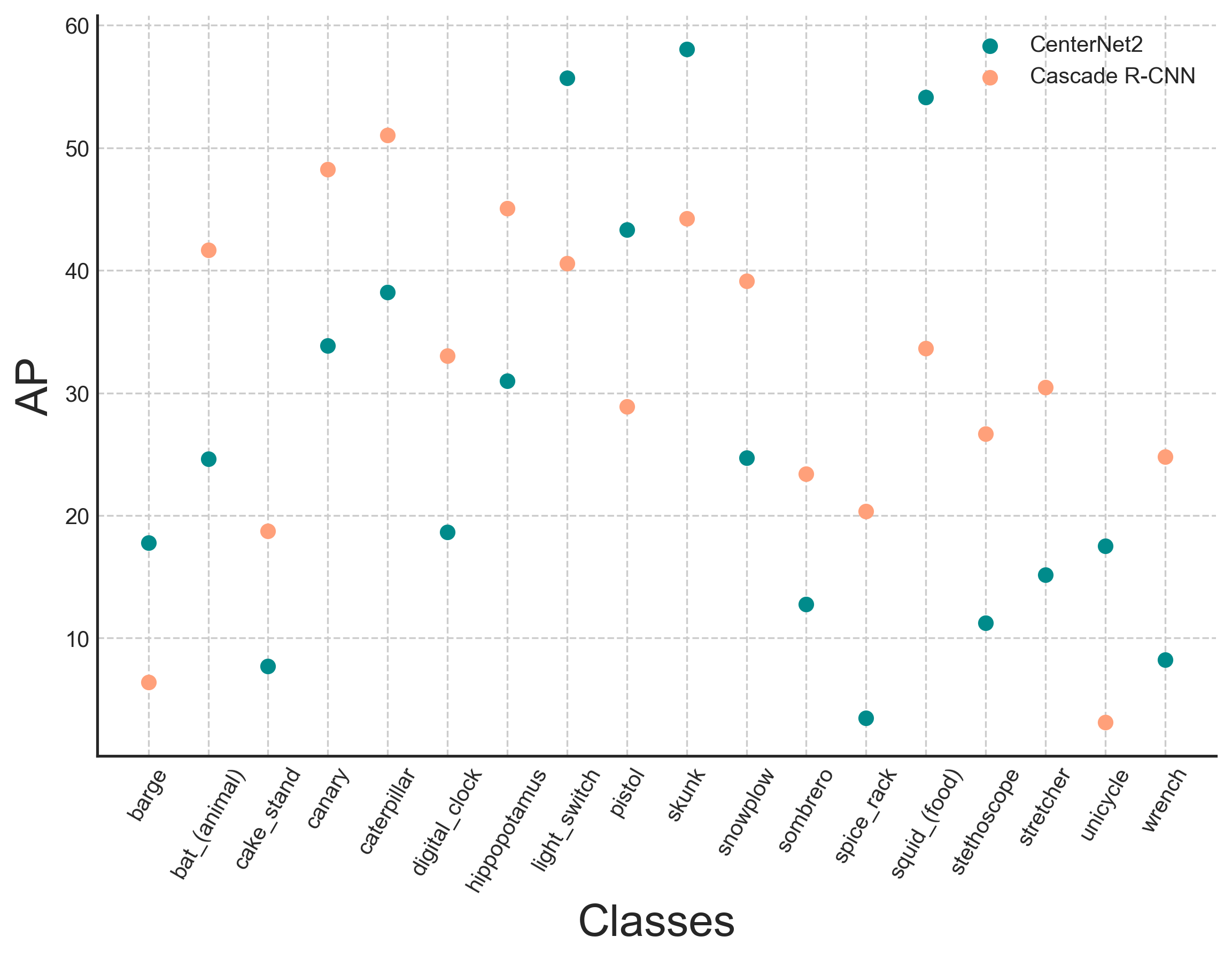}
  \caption{Comparison of AP results between different teacher models on several classes.}
  \label{fig:ap_gap}
\end{figure}

We provide the implementation details of pseudo annotation generation mechanism. 
As mentioned in \cref{subsec:self-training}, we adopt a multi-teacher strategy to improve the credibility of pseudo annotations. 
There are two teacher models utilized in our work as shown in \cref{tab:multi-teacher}. For both teachers, we set the score threshold as $0.5$ and the NMS threshold as $0.6$. 
After training, two teacher models achieve $23.1$ and $24.1$ AP on \verb+bigdet_val+ respectively, which are quite close. 
In \cref{fig:ap_gap}, we show the comparison of AP results between two teacher models on single class. 
It can be seen that two teacher models have significant difference on single class, and the results gap even reaches $20$ AP for class ``\textit{squid\_(food)}''. 
Mixing the outputs of multiple teacher models will fix the AP gap on single class, thereby improving the quality of pseudo annotations.

\begin{table}[ht]
    \begin{center}
        \begin{tabular}{c|cccc}
            \toprule
             & Model & Schedule & Score & NMS \\
            \midrule
            T1 & CenterNet2 & 8$\times$ & 0.5 & 0.6 \\
            T2 & Cascade R-CNN & 8$\times$ & 0.5 & 0.6 \\
            \bottomrule
        \end{tabular}
    \end{center}
    \caption{Basic setting for two teacher models. For both models, an $8\times$ training schedule is adopted, and score and NMS threshold are set as $0.5$ and $0.6$.}
    \label{tab:multi-teacher}
\end{table}

\section{Generalization Ability}
\label{sec:general}

One of the great benefits of pre-training on a large-scale dataset is a well-trained model only needs a few target labels to perform considerably well. Here, we show BigDetection pre-training is helpful across a variety of dataset sizes and semantic domains, and helps data efficiency.

Following the partially labeled data setting mentioned in paper, CenterNet2~\cite{zhou2021probabilistic} with FPN is adopted for fair comparison. The finetuning is done on PASCAL VOC~\cite{everingham2015pascal} and Cityscapes~\cite{cordts2016cityscapes} using $1\%$ and $5\%$ samples of train split. 
\cref{tab:multi-data-effi} compares the results of ImageNet~\cite{deng2009imagenet} pre-trained model (Supervised), OpenImages~\cite{kuznetsova2020open} pre-trained model and BigDetection pre-trained model. 
Comparing with existing largest detection pre-training dataset, model pre-trained on BigDetection has better performance when dealing with insufficient training data. 

\begin{table}[ht]
    \begin{center}
        \begin{tabular}{c|cc|cc}
            \toprule
            \multirow{2}*{Methods} & \multicolumn{2}{c|}{VOC} & \multicolumn{2}{c}{Cityscapes} \\
            ~ & $1\%$ & $5\%$ & $1\%$ & $5\%$ \\
            \midrule
            ImageNet & 23.4 & 50.2 & 12.4 & 24.3  \\
            OpenImages & 58.3 & 67.1 & 23.3 & 30.1 \\
            \midrule
            BigDetection & \textbf{64.6} & \textbf{72.6} & \textbf{31.8} & \textbf{38.9} \\
            \bottomrule
        \end{tabular}
    \end{center}
    \caption{Comparing with different pre-trained models under multiple partially labeled datasets.}
    \label{tab:multi-data-effi}
    \vspace{-2ex}
\end{table}

\section{Qualitative Results}
\label{sec:qr}

\begin{figure*}[t]
    \centering
    \subfloat[OpenImages results]{
        \includegraphics[width=0.48\linewidth]{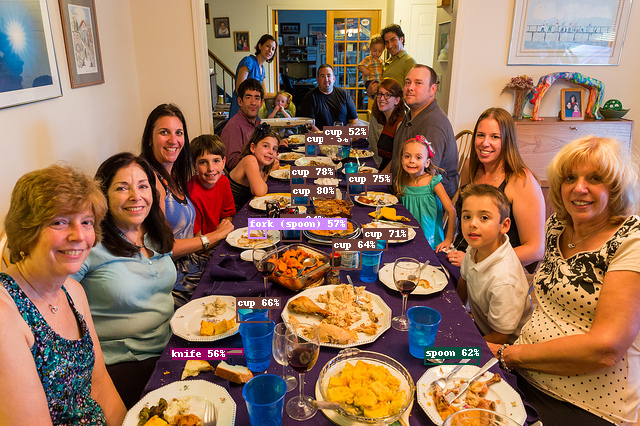}
        \label{fig:qr_1_oid}
    }
    \subfloat[BigDetection results.]{
        \includegraphics[width=0.48\linewidth]{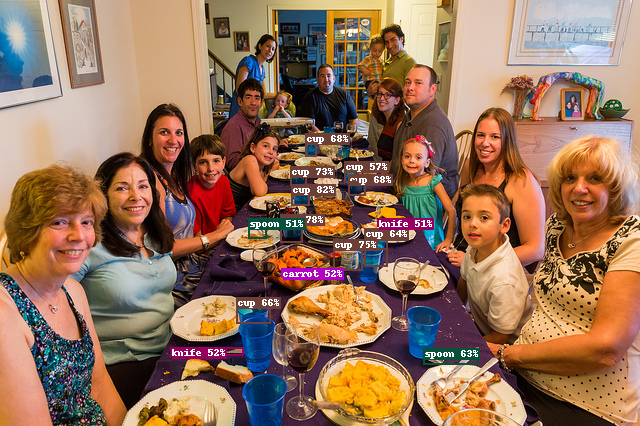}
        \label{fig:qr_1_bigdet}
    }
    
    \subfloat[OpenImages results]{
        \includegraphics[width=0.48\linewidth]{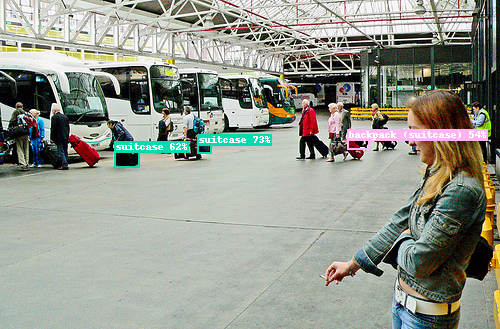}
        \label{fig:qr_2_oid}
    }
    \subfloat[BigDetection results]{
        \includegraphics[width=0.48\linewidth]{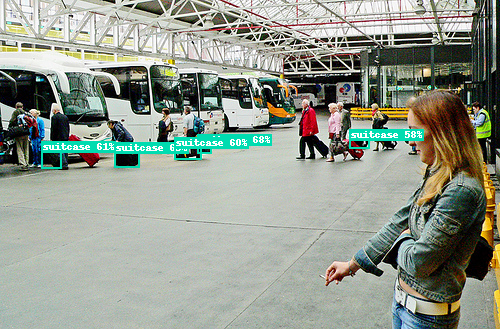}
        \label{fig:qr_2_bigdet}
    }
    
    \subfloat[OpenImages results]{
        \includegraphics[width=0.48\linewidth]{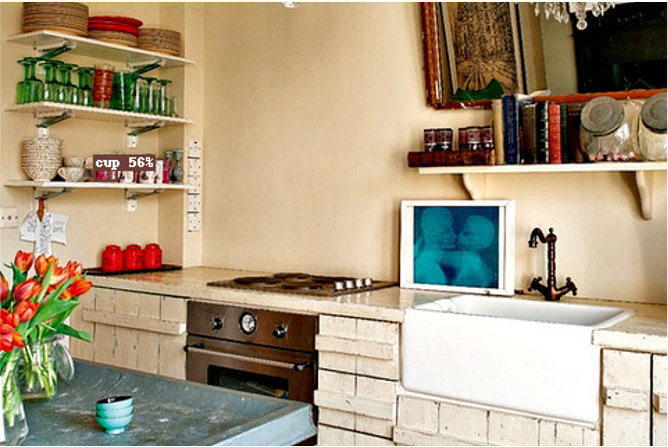}
        \label{fig:qr_3_oid}
    }
    \subfloat[BigDetection results]{
        \includegraphics[width=0.48\linewidth]{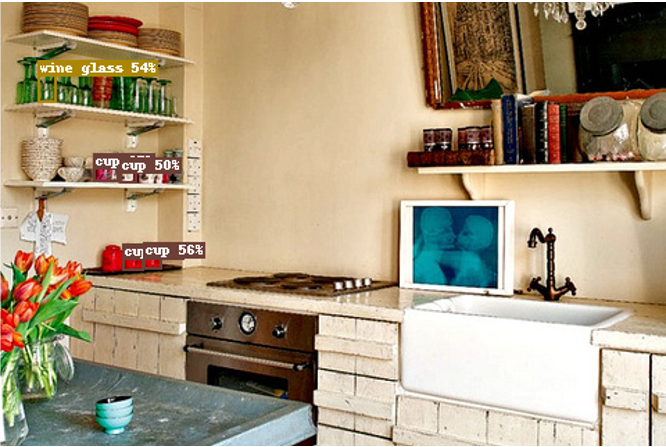}
        \label{fig:qr_3_bigdet}
    }
    
    \caption{Visual comparison on results of OpenImages and BigDetection pre-trained models. Left column shows detection results of OpenImages. Right column shows detection results of BigDetection. 
    First row: ``\textit{knife}'', ``\textit{carrot}'' and ``\textit{cup}'' classes have several small-scale objects that are not captured by OpenImages pre-trained model, even ``\textit{spoon}'' is misclassified as ``\textit{fork}''. Most of these objects present in the detection results of BigDetection pre-trained model with correct label predictions. 
    Second row: results of small-scale object ``\textit{suitcase}''. These objects are either not detected, or are misclassified by OpenImages pre-trained model, while almost all of them are captured by BigDetection pre-trained model. 
    Third row: results of classes ``\textit{cup}'' and ``\textit{wine glass}''. BigDetection pre-trained model can better capture these small-scale objects while OpenImages pretrained model cannot.}
    \label{fig:qr}
\end{figure*}

In this part, we visualize detection results on images from COCO validation set that contain multiple small-scale objects. CenterNet2~\cite{zhou2021probabilistic} pre-trained on OpenImages~\cite{kuznetsova2020open} and BigDetection with $8\times$ schedule are adopted for comparison. 
As shown in \cref{fig:qr}, left column shows detection results of OpenImages pre-trained model, and right column shows results of BigDetection pre-trained model. 

In first row (\cref{fig:qr_1_oid} and \cref{fig:qr_1_bigdet}), for ``\textit{knife}'', ``\textit{carrot}'' and ``\textit{cup}'' classes, several small-scale objects are not captured by OpenImages pre-trained model, even ``\textit{spoon}'' is misclassified as ``\textit{fork}''.
However, most of these objects present in the detection results of BigDetection pre-trained model with correct class predictions.
In second row (\cref{fig:qr_2_oid} and \cref{fig:qr_2_bigdet}), detection results on small-scale object ``\textit{suitcase}'' is illustrated. These objects are either not detected, or are misclassified by OpenImages pre-trained model, while almost all of them are captured by our system.
Finally, in third row (\cref{fig:qr_3_oid} and \cref{fig:qr_3_bigdet}), results of classes ``\textit{cup}'' and ``\textit{wine glass}'' are shown. BigDetection pre-trained model can better capture these small-scale objects while OpenImages pre-trained model cannot.
In summary, BigDetection pre-trained model has better performance when facing small-scale objects, including capturing more small-scale objects and more accurate class prediction.

\section{Synsets of BigDetection}
\label{sec:synsets}

In addition, we provide the final synsets of BigDetection dataset, which are obtained by our manual data-cleaning and careful designed category mapping principles. The synsets file \textit{bigdetection\_synsets.txt} is included in: \url{https://github.com/amazon-research/bigdetection}.

\end{document}